\documentclass[letterpaper, 10 pt, conference]{ieeeconf}  

\IEEEoverridecommandlockouts                              

\usepackage{amsmath} 
\usepackage{amssymb}  
\usepackage{bm}
\usepackage{graphicx}
\usepackage{booktabs}
\usepackage{xcolor}
\usepackage{algorithm}
\usepackage{algpseudocode}
\usepackage{comment}
\usepackage{float}
\usepackage{relsize}

\title{\LARGE \bf
Learning to Imitate Spatial Organization in Multi-robot Systems
}

\author{Ayomide O. Agunloye$^{1}$, Sarvapali D. Ramchurn$^{1}$, and Mohammad D. Soorati$^{1}$
\thanks{$^{1}$Authors are with the School of Electronics and Computer Science, University
of Southampton, Southampton, SO17 1TR, United Kingdom.
        {\tt\small \{a.o.agunloye, sdr1, m.soorati\} @soton.ac.uk}}%
}

\begin{document}

\maketitle
\thispagestyle{empty}
\pagestyle{empty}

\begin{abstract}
Understanding collective behavior and how it evolves is important to ensure that robot swarms can be trusted in a shared environment. One way to understand the behavior of the swarm is through collective behavior reconstruction using prior demonstrations. Existing approaches often require access to the swarm controller which may not be available. We reconstruct collective behaviors in distinct swarm scenarios involving shared environments without using swarm controller information. We achieve this by transforming prior demonstrations into features that describe multi-agent interactions before behavior reconstruction with multi-agent generative adversarial imitation learning (MA-GAIL). We show that our approach outperforms existing algorithms in spatial organization, and can be used to observe and reconstruct a swarm's behavior for further analysis and testing, which might be impractical or undesirable on the original robot swarm.

\end{abstract}

\section{INTRODUCTION}
Swarm robotics and its applications are transitioning from the laboratory to the real world \cite{dorigo_swarm_2021, cheraghi_past_2022}, and it is expected to lead to the large-scale deployment of multiple robots in environments that are shared between robots and humans. For the robot-robot and human-robot interactions to be seamless, robot swarms must be safe and trustworthy \cite{soorati202493}. Ensuring that a swarm is safe and trustworthy in shared environments requires precise and continuous knowledge of their collective behavior and how it evolves. In scenarios where swarm controllers are directly accessible, collective behavior can be modeled using the controllers \cite{ligot_using_2022}. However, in practical scenarios, swarm controllers may not be accessible for various reasons (e.g., inability to extract controllers from natural swarms).  In robot swarms, access to controllers may be restricted or impractical due to various reasons such as encryption of controller information due to strategic or privacy concerns \cite{hunt_demonstrating_2023, chen_privacy-preserving_2022}. To this end, understanding the collective behavior of a swarm and modeling its dynamics without using swarm controller information requires thorough research.

Collective behavior reconstruction and recognition are two established methods for modeling swarm dynamics or explaining collective behavior \cite{naiseh2022}. Collective behavior reconstruction can be model-based or data-driven. In model-based approaches, behavior reconstruction is achieved using a mathematical or regression model \cite{reynolds_flocks_1987, sinhuber_equation_2021}. In data-driven approaches, multi-agent interactions are statistically extracted or learned from prior demonstrations to reproduce observed behavior. Recently, data-driven approaches used imitation learning (IL) algorithms such as inverse reinforcement learning (IRL) and generative adversarial imitation learning (GAIL) for improved reconstruction accuracy in swarm scenarios modeled as multi-agent systems \cite{waelchli_discovering_2023,sosic_inverse_2017,pinsler_inverse_2018, costa_automated_2020, liu_pbt-gail_2022, yu_swarm_2021}. Genetic programming and graphical neural networks have also been used for data-driven behavior reconstruction with swarm controllers extracted from video demonstrations \cite{alharthi_automatic_2022} and swarm behavior prediction \cite{zhou_clone_2019}. For recognition of collective behaviour, binary classification of observed behavior as defined or undefined collective behavior is a common approach \cite{khan_autonomous_2020, khattab_autonomous_2023, li_turing_2016}. This approach does not scale, as each swarm scenario requires a unique classifier. Multinomial classification addresses this and has been achieved in closely related swarming scenarios using predefined and learned multi-agent interaction quantifiers \cite{cenek_towards_2016, brown_limited_2014,abpeikar_automatic_2023}.

Existing data-driven behavior reconstruction approaches extract multi-agent interactions from expert demonstrations without capturing swarm-environment interactions. As a result, the recovered multi-agent interactions cannot accurately reconstruct or predict expert behavior. While this issue is addressed in~\cite{zhou_clone_2019}, their approach relied on learned extraction of multi-agent interactions and it is difficult to explain how the robots interact with the environment.

In this work, we investigate the reconstruction of collective behavior in three spatial organization tasks involving shared environments without using swarm controller information. We consider three common swarm robotic scenarios: aggregation, homing, and obstacle avoidance. We model these scenarios as single-objective swarm scenarios where swarming agents interact with each other and the environment while completing the swarm objective. We generate informed and explainable multi-agent interactions through feature transformation of expert demonstrations and, achieve near-optimal behavior reconstruction using multi-agent GAIL. We show that even when learner robots are initialized from unforeseen states, they perform similarly to the expert robots in all investigated scenarios.

The main contributions of this paper are as follows. (1) We present an approach for reconstructing collective behaviors in shared environments without accessing robot controllers; and (2) We demonstrate the use of informed and explainable multi-agent interactions for improved learning representation in data-driven behavior reconstruction.

\section{RELATED WORKS}
IRL has been extensively used in the literature for the reconstruction of collective behaviour as it recovers the underlying reward functions while reproducing expert behavior. Šošić et al.~\cite{sosic_inverse_2017} reconstructed the behavior of a homogeneous swarm by assuming that all agents are interchangeable and share a central reward function, thereby reducing the problem to a single-agent IRL. Another study~\cite{pinsler_inverse_2018} reconstructed the collective behavior observed in a flock of pigeons by recovering individual reward functions for each pigeon. This individualistic approach exposed the multi-agent interactions in the flock and allowed the researchers to model the leader-follower hierarchy. In \cite{gharbi_show_2023}, a similar individualistic reward function approach was used to evolve the robot controller using IRL by manually specifying the desired goal location or the path. Other studies~\cite{costa_automated_2020,waelchli_discovering_2023} extended the IRL framework to multi-agent IRL to simultaneously recover the reward functions of several agents. Liu et al.~\cite{liu_pbt-gail_2022}, however, integrated GAIL with population-based training for collective behavior reconstruction. Besides IRL and GAIL, other machine learning methods have been used to reconstruct collective behavior. In \cite{alharthi_automatic_2022}, genetic programming was used to extract explainable controllers from video demonstrations of collective behavior with $8$ predefined swarm interaction metrics defining the fitness measure. Zhou et al.~\cite{zhou_clone_2019} used graphical neural networks to imitate the behavior of expert robots and predict trajectories. Their approach considers swarm-environment interactions and modeled robots and environmental entities as graph nodes, but does not provide explainable interactions or controllers. Most of these works demonstrate behavior reconstruction in a single scenario or multiple similar scenarios (e.g., swarming and schooling). However, Yu et al.~\cite{yu_swarm_2021} used Adversarial Imitation Learning with parameter sharing (PS-AIRL) for behavior reconstruction in distinct swarming scenarios. Their approach focused on homogeneous biological swarms and did not consider swarm-environment interactions. They also require access to the original swarm controllers which is rarely available in practical scenarios. In contrast, our approach considers swarm-environment interactions and reconstructs expert behavior without accessing robot controllers. We also generate informed multi-agent interactions that can be used to explain swarm behavior. 

\section{BACKGROUND}   
\label{sec:background}
We consider decentralized Partially Observable Markov Decision Processes (Dec-POMDP)~\cite{oliehoek_concise_2016} in which agents receive individual rewards for their actions. A Dec-POMDP is defined as an MDP comprising a tuple $\langle \mathcal{N}, \mathcal{S}, \mathcal{A}, \mathcal{T}, \mathcal{R}, \mathcal{O}, \Omega, \gamma \rangle$. $\mathcal{N}$ represents the set of agents in the Dec-POMDP, $\mathcal{S}$ is the global state space of the environment, $\mathcal{A}$ contains the shared action space of all agents in $\mathcal{N}$ and $\mathcal{O}$ represents the joint observation space of all agents in the environment. At each episodic time step \textit{t}, each agent $i \in \mathcal{N} \equiv \{ 1, ..., n \}$ takes an action $a_i \in \mathcal{A}$ to form the joint action $\textbf{\textit{a}} \in \bm{\mathcal{A}} \equiv \mathcal{A}^n$ based on its partial observation of the environment $o_i \in \Omega$ as provided by the observation function $\mathcal{O}(s,a)$ using parameterized policy $\pi_i(a_i|o_i)$. The state transition function $\mathcal{T}(s\prime|s,\textbf{\textit{a}}): \mathcal{S} \times \bm{\mathcal{A}} \to \mathcal{S}$ provides the next global state, and the shared reward function $r(s, \textbf{\textit{a}}) : \mathcal{S} \times \bm{\mathcal{A}} \to \mathcal{R}$ gives each agent an individual reward $r_i \in \mathcal{R}$. $\gamma \in [\,0,1)$ denotes the reward discount factor.

GAIL achieves imitation learning by matching the occupancy measures $\rho_{\pi_E}$ of the expert policy $\pi_E$ in the learner domain through generative adversarial training \cite{gui_review_2023}. The occupancy measure is the unnormalized distribution of an agent's trajectory as it navigates the environment using a policy $\pi$ \cite{ho_generative_2016}. In GAIL, the generator is a policy network $\pi$ that produces trajectories from a similar environment as the expert. The discriminator network $D$ compares generated trajectories with expert demonstrations and attempts to distinguish them through binary classification.

The GAIL objective function can be written in terms of occupancy measures and expectations over expert and learner policies as \cite{ho_generative_2016}:
\begin{multline}
    \psi^{\star}_{GA} (\rho_{\pi} - \rho_{\pi_E})
    = \underset{D \in (0,1)^{\mathcal{S} \times \mathcal{A}}}{\text{max}} \mathbb{E}_{\pi} [\log (D(s,a))] + \\
    \mathbb{E}_{\pi_E} [\log (1 - D(s,a))]
    \label{eq:gail}
\end{multline}
where $\psi^{\star}_{GA}$ is the convex regularization imposed on the generator by $D$, $D(s,a)$ is the discriminator output, and $\log D(s,a)$ is the learning signal for the generator.

GAIL optimizes Equation \ref{eq:gail} by finding its saddle point $(\pi, D)$. At this point, $D$ is unable to differentiate between trajectories from $\pi$ and $\pi_E$. When $\pi$ and $D$ are represented by function approximators, GAIL fits a parameterized policy $\pi_\theta$ and a discriminator network $D_{w}$ with weights $w$. The discriminator feedback serves a reward function that encourages the generator to minimize the dissimilarity between $\rho_{\pi}$ and $\rho_{\pi_E}$.
    
In multi-agent systems, individual agents optimize separate reward functions that describe their behavior. As a result, multiple reward functions exist and optimality is only guaranteed through a set of stationary policies that provide a Nash equilibrium solution. Multi-agent GAIL addresses this by jointly optimizing the Nash equilibrium constraints with the objective function during occupancy measures matching \cite{song_multi-agent_2018}.

\section{METHOD}
In this section, we describe our approach to accurate collective behavior reconstruction. We formulate the problem as a collective behavior reconstruction problem in a shared Dec-POMDP environment. Expert demonstrations $\mathcal{D}$ contain the absolute position of all observable entities $\mathcal{M}$ in the environment. We transform $\mathcal{D}$ into informed multi-agent interactions before recovering policies that accurately reproduce expert behaviors using multi-agent GAIL (MA-GAIL)~\cite{song_multi-agent_2018}. 

\subsection{Expert Demonstrations Transformation}
We transform each expert trajectory in $\mathcal{D}$ to a set of state representative features $\mathbf{f}_s$ describing the interaction between the expert and all other observable entities in the environment given a state $s \in \mathcal{D}$. We achieve the transformation by computing the cohesion between agent $i$ and every other entity in $\mathcal{M}$. Thus, the state representative features for agent $i$ in state $s$ is:
\begin{equation}
    \mathbf{f}^{i}_{s} = [-dist(i,j) | j \in \mathcal{M}, j \neq i]
    \label{eq:feature}
\end{equation}
where $dist(i,j)$ denotes the euclidean distance between agent $i$ and entity $j$.
    
\subsection{Policy Recovery with Multi-Agent GAIL}
To recover stationary policies in the DEC-POMDP, we use MA-GAIL with $n$ individual discriminators $\mathbf{D}= \{ D_1, D_2, ..., D_n\}$ and match occupancy measures on transformed expert demonstrations. For the generator network ${\pi}$, we use Multi-Agent Proximal Policy Optimization (MAPPO) with parameter sharing in which all learners use a single policy network. This applies to our environment since our agents are homogeneous and have identical observation and action spaces \cite{yu_surprising_2022}. Using individual discriminators ensures that each learner strictly matches the occupancy measures of a particular expert. However, this prevents generalization for homogeneous agents as learners receive poor feedback if they behave like any other expert. We address this through expert demonstration sharing and allow individual discriminators to compare trajectories from their learners with all expert demonstrations available. This ensures that learners are positively rewarded for demonstrating any valid expert behavior instead of the particular behavior from one expert.

The policy recovery algorithm is summarized in Algorithm~\ref{alg:policy_recovery}. Given expert demonstrations $\mathcal{D}$, learners interact with the environment and generate rollout trajectories $T_k$. The discriminators are trained using feature transformed $\mathcal{D}$ and $T_k$. At each time step, learners receive individual reward feedback $r_{\pi, D_n}$ with which the shared policy is improved. Compared to PS-AIRL \cite{yu_swarm_2021}, our algorithm uses $n$ discriminators instead of one and transforms all input into state representative features before passing them to the discriminators. It also allows the discriminators to share the features for improved learning representation.

\begin{algorithm}
\caption{Policy recovery with MA-GAIL}\label{alg:policy_recovery}
\begin{algorithmic}
\State Input: expert demonstrations $\mathcal{D}$
\State Randomly initialize generator ${\pi}$ \& discriminators $\mathbf{D}$
\For {$k = 1,2, \ldots $}
\State Rollout learner $\bm{T} = \{T_1, T_2, ..., T_k\}$ using ${\pi}$
\For {$n = 1,2, \ldots, |\mathcal{N}|$}
\State {Train $D_n$ to classify $\mathbf{f}^{n}_{s} \forall s \in \mathcal{D}$ from $\mathbf{f}^{n}_{s} \forall s \in T_k$}
\EndFor
\State Generate $r_{\pi, D_n}$ for each generator policy
$$ r_{\pi, D_n} \gets [\log (D_n(\mathbf{f}^{n}_{s}))] + [\log (1 - D_n(\mathbf{f}^{n}_{s})] $$
\State Update $\pi$ using $r_{\pi, D_n}$ with PS-MAPPO
\EndFor
\end{algorithmic}
\end{algorithm}

\section{EXPERIMENT}
We evaluate the performance of our proposed approach in three classical swarm robotic scenarios: aggregation, homing, and obstacle avoidance. We model these scenarios as cooperative and single-objective in a shared environment. We consider a swarm size of $3$ and represent our swarming agents as uncrewed aerial vehicles (UAVs) with inaccessible controllers. To improve learning representation, we reduce the complexity of the shared environment and separate it into motion and control layers. The control layer is a discretized grid world representation of the shared environment with reduced state and action space. We compare the performance of the proposed approach with  PS-AIRL \cite{yu_swarm_2021} and behavior cloning (BC), where a direct mapping between expert states and actions is learned \cite{zheng_imitation_2022}.

\subsection{Swarm Scenarios}
        

In \textit{aggregation}, the objective of swarming UAVs is to maximize the intra-swarm cohesion. They achieve this by safely forming a cluster at any suitable zone in a shared environment. The shared environment includes two active UAVs hovering at fixed positions. Swarming UAVs can observe fixed-position UAVs if they are within perception range in both layers but can only interact with them in the motion layer. We model the individual reward at each time step in the control layer $r_n$ as: 
\begin{equation}
r_n = 
\begin{cases} 
    n_{\text{agents}} \times \text{c} & \text{if } n_{\text{agents}} > 1 \\
    -\text{c} & \text{otherwise}
\end{cases}
\label{eq:agg_reward}
\end{equation}
where $n_{\text{agents}} = |\{c_n > t \forall n \in \mathcal{N}\}|$, and $t$ is an environment specific aggregation threshold.

The objective in the \textit{homing} scenario differs from aggregation in that the clustering zone---home position---is fixed and cannot be dynamically chosen by the robots. In this scenario, the UAVs must explore the environment and locate the home positions before the episode ends. Once a UAV finds a home position, it must remain there until all other UAVs have~\textit{homed}. We model $r_n$ as the maximum cohesion between UAV $n$ and any home position at a given time step.
    
In a new behavior that we refer to as \textit{obstacle avoidance}, UAVs must navigate the shared environment without interacting with fixed-location inactive UAVs in the shared environment. This scenario differs from existing obstacle avoidance scenarios in that the UAVs can access the positions already occupied by the inactive UAVs in the control layers. However, they receive a large negative reward for doing this. The motivation for this behavior is that it is crucial to maintain a safe distance from unknown entities in a practical shared environment, even if they seem inactive. The UAVs also receive a small negative reward for insufficient exploration. We model $r_n$ as a large constant $-c$ when the cohesion between the UAV $n$ and any fixed position UAV is maximized and $0$ otherwise.

\begin{figure}[hbt!]
\centering
\includegraphics[width=\columnwidth]{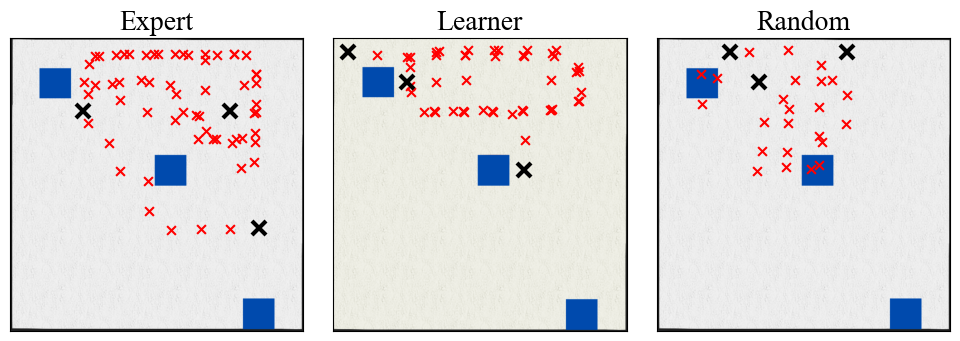}     
\caption{Snapshot of the motion layer showing the obstacle avoidance behavior and position trace for Experts, Learners, and Random swarming UAVs between t=$0$ and t=$300$s. \textbf{X} represents UAVs positions at t=$0$. Blue boxes are inactive UAVs locations.} 
\label{fig1}
\end{figure}
\subsection{Simulation Environment}
We implement the motion layer for the shared environment in Webots \cite{webots_cyberbotics_nodate} using simulated Crazyflies 2.0 drones \cite{giernacki_crazyflie_2017} as the swarm. The simulation boundary is a $3$m by $3$m continuous rectangular world. UAVs can move in all directions and can detect obstacles using onboard range sensors.

The control layer is a $10 \times 10$ grid world environment. Here, the action space in the is limited to $a \in \mathbb{R}^{5}$ corresponding only to the high-level control of the UAVs \textit{\{stop, right, left, forward, and backward\}}. Low-level motion controls such as lift-off, turning, translation, and hovering are implemented deterministically in the motion layer. Agents in the grid world can observe the positions of other entities up to $6$ grid positions in all directions. Given that only two fixed-position entities are in the aggregation scenario, the agent's observation space is $o \in \mathbb{R}^{10}$ in this scenario and $o \in \mathbb{R}^{12}$ in others. All episodes run for a fixed duration of $50$ time steps in the control layer. This corresponds to an episode duration of about $300s$ in the motion layer. Figure \ref{fig1} shows the position trace of the experts, learners, and random UAV behavior in the obstacle avoidance scenario between $t=0$ and $t=300s$ with small red crosses. The three larger crosses on each setup mark the initial positions.

\subsection{Implementation Details}
\subsubsection{Expert Demonstrations}
To generate expert demonstrations of collective behavior in each scenario, we train expert UAVs in the corresponding Dec-POMDP grid world using PS-MAPPO for $100,000$ training episodes. After training, we generate an expert demonstration data pool of $1,000$ trajectories in each scenario. We also generate noisy expert demonstrations by varying expert optimality $\epsilon \in [0,1]$, where $0$ implies optimal experts and $1$ implies experts sampling actions at random. It should be noted that trajectories in the expert demonstrations data pool are randomly generated and may contain similar expert UAV behavior.

\subsubsection{Learner UAVs}
Learner UAVs interact with the environment for $10,000$ training episodes using expert demonstrations between $200$ and $500$. This expert demonstration range was chosen as it agrees with expert dataset sizes in existing works \cite{zhou_clone_2019, yu_swarm_2021}. Learner UAVs receive individual rewards from their discriminator for each episodic time step. The rewards and corresponding trajectories are stored in a shared buffer for training the PS-MAPPO policy at the end of each episode. After $50$ training episodes, the discriminators are first initialized and trained using available learner and expert trajectories. They are then updated every $50$th episode for $1,000$ training episodes and then every $500$th episode. This update frequency ensures that the discriminators are properly initialized but do not change too quickly, thus allowing learner UAVs to understand reward patterns.

\subsubsection{Network Training}
All models were trained and evaluated on a single cluster node with a 64 cores 2.2 GHz Intel CPU and 256 GB of RAM. Expert policies training took about $7$ hours per scenario, while learner policies training only took an hour per scenario. PS-MAPPO implementations for expert policy and MA-GAIL generator network used the default hyperparameters provided in the original paper \cite{yu_surprising_2022}. The MA-GAIL discriminators were simple $2$ layer multi-layer perceptron network (MLP) with $128$ hidden units and rectified linear unit (relu) activations. These discriminators were trained in parallel using a learning rate of $1 \times 10^{-5}$ so that their training does not influence the training time. PS-AIRL was implemented using the algorithm provided in the paper while BC was achieved using individual $3$ layer MLPs with $128$ hidden units and relu activation.

\begin{figure}[!htpb]
\centering
    \includegraphics[width=\columnwidth]{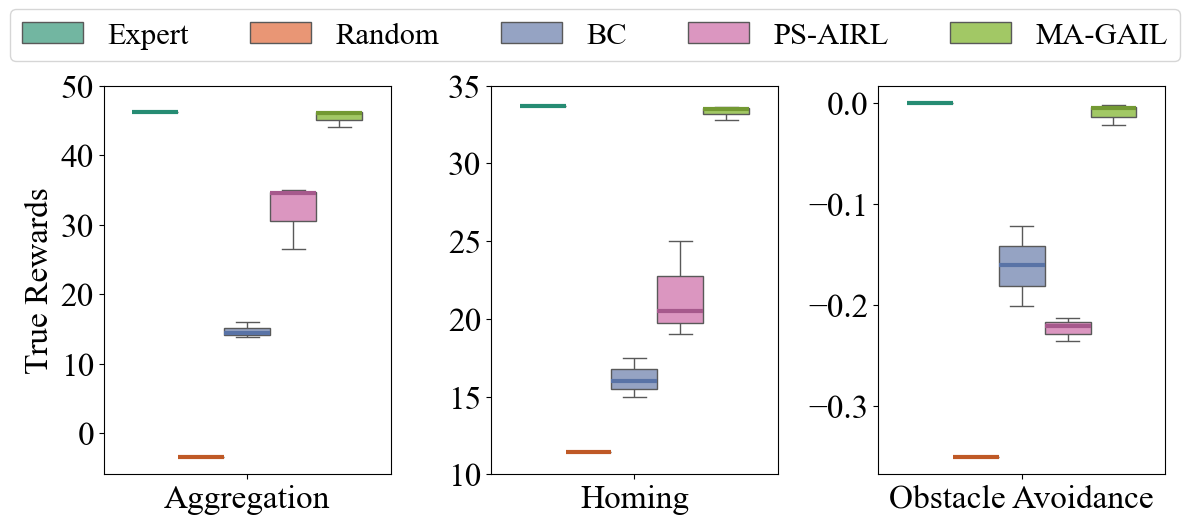}
    \caption{Boxplots of true episode rewards obtained in $200$ evaluation episodes by the proposed approach (MA-GAIL), BC, and PS-AIRL trained with $400$ expert demonstrations in all scenarios.}
    \label{fig2}
\end{figure}
\section{RESULTS}
We demonstrate the advantage of our behavior reconstruction algorithm in two different ways. In the first setup, we initialize learner UAVs from starting positions present in $\mathcal{D}$ for every evaluation episode. Figure~\ref{fig2} shows the performance comparison between the proposed approach (MA-GAIL), PS-AIRL, and BC trained using $400$ expert demonstrations over $200$ evaluation episodes with unnormalized reward values in all scenarios. As the figure demonstrates, our approach closely reproduces expert behavior in all scenarios compared to BC and PS-AIRL. This high performance across distinct swarm scenarios can be attributed to the transformed expert demonstrations, which sufficiently describe the multi-agent interactions in the shared environment. It can also be attributed to expert demonstration sharing, which increases the set of valid expert behaviors, thus allowing learners to easily reproduce expert behaviors regardless of the scenario. PS-AIRL outperformed BC in aggregation and homing scenarios but failed to maintain its superiority in the obstacle avoidance scenario. We attribute this to the sparsity of the reward function in the obstacle avoidance scenario, which forces the experts to conservatively explore a small area in the shared environment and avoid the large negative rewards. The abundance of sequential data from this region makes it easy for BC to \textit{clone} expert actions when initialized from starting positions close to the area and outperform PS-AIRL. Conversely, the continuous reward function in aggregation and homing scenarios provides a normal distribution of state-action pairs in expert demonstrations making it difficult for BC to \textit{clone} expert actions. It should be noted that evaluation results are from learners trained using $400$ expert demonstrations as they represent the best-performance region for PS-AIRL and BC.  

\begin{figure}[!htpb]
\centering
    \includegraphics[width=0.95\columnwidth]{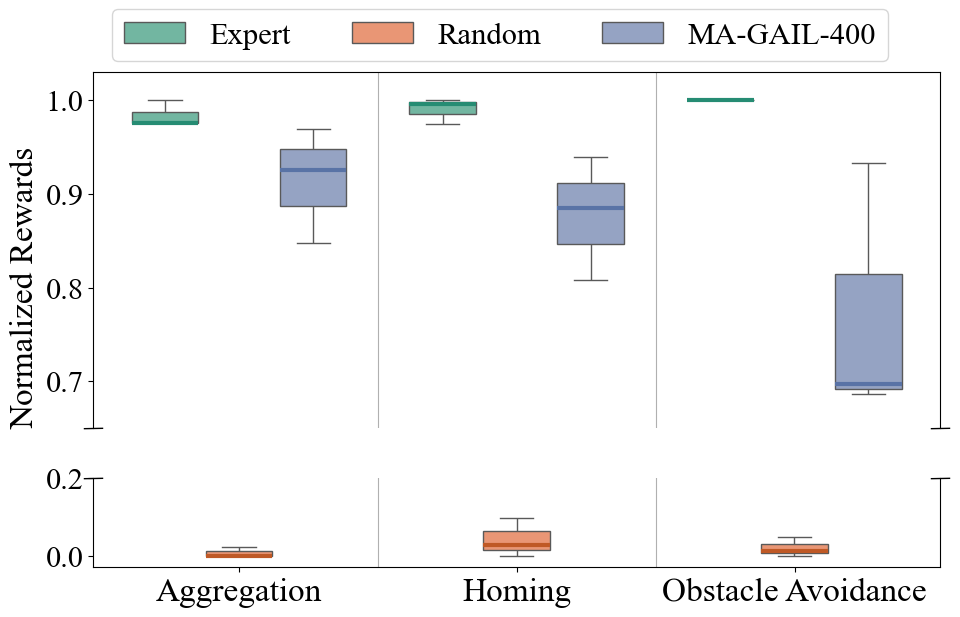}
    \caption{Boxplots of normalized reward values for Expert, Random, and MA-GAIL-400 over $200$ evaluation episodes initialized from random starting states in all scenarios.}
    \label{fig3}
\end{figure}

\begin{figure}[!htpb]
    \centering
        \includegraphics[width=\columnwidth]{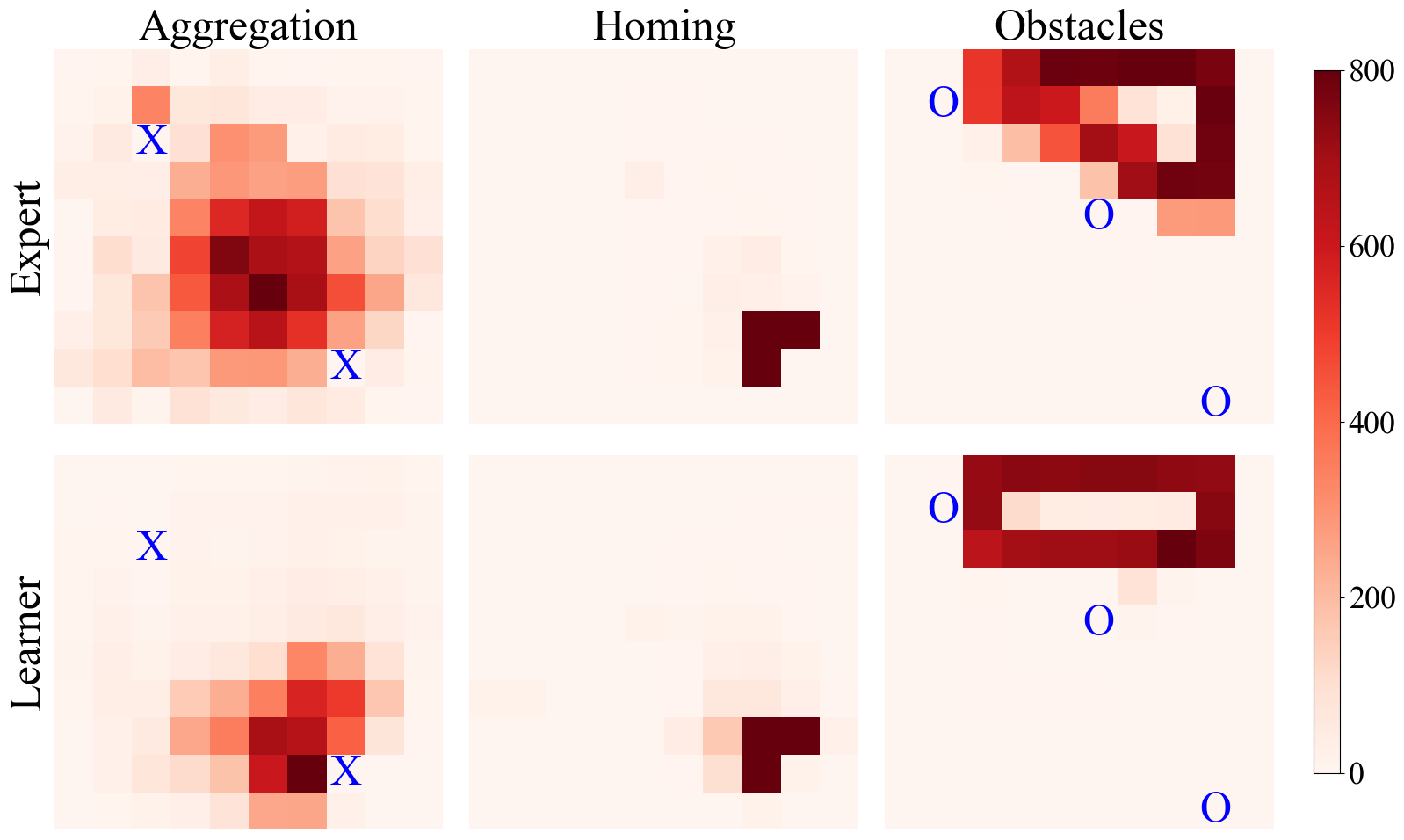}
        \caption{Visualization of swarming UAVs positions for $10$ evaluation episodes in all scenarios. Area coverage of expert (top) and learner (bottom). Active UAV locations are marked by \textbf{\textcolor{blue}{{X}}} and inactive locations are shown as \textbf{\textcolor{blue}{{O}}}.}
        \label{fig4}
    \end{figure}
In the second evaluation setup, learner UAVs are initialized randomly from unforeseen starting states at the beginning of each evaluation episode.  Figure~\ref{fig3} presents the normalized reward values for the experts, random (suboptimal experts with $\epsilon = 1$), and learners trained with $400$ expert demonstrations (MA-GAIL-400) over $200$ different evaluation episodes in all scenarios. As the figure demonstrates, learners do not perfectly reproduce expert behaviors in all scenarios due to their non-familiarity with the initial states. This effect is, however, pronounced in the obstacle avoidance scenario where expert agents can safely navigate the fixed-position entities without interacting with them, even though it is~\textit{risky} due to the sparsity of the reward function. Learner UAVs, on the other hand, do not consistently reproduce this~\textit{risky} behavior when initialized from random starting states. This shows that while imitating the controllers of multi-robot systems generated with a sparse reward or cost function may be easy, accurately predicting how they will perform in unforeseen states still requires further research. It should be noted that modeling the obstacle avoidance scenario using a continuous cohesion-based reward function did not produce optimal expert policies in the control layer.

Figure~\ref{fig4} shows the area coverage of optimal experts and MA-GAIL-400 learner UAVs in $10$ evaluation episodes in all scenarios. The fixed-position active UAVs are marked with `X' in aggregation, while inactive UAVs are represented as `O' in obstacle avoidance. We observe that learner UAVs do not directly reproduce particular expert behavior but unravel patterns in the demonstrations that allow them to maximize discriminator reward and mimic any expert. This is evident in the aggregation and obstacle avoidance scenarios where learner UAVs do not explore the shared environment as much as the experts, even though they are initialized from the same starting positions. Variations in expert and learner UAVs' absolute positions in Figure \ref{fig4} result from learner UAVs matching the occupancy measures of features describing expert behaviors in the control layer and not their absolute positions. This is intuitive since the GAIL convex regularizer only penalizes the generator heavily when it maximizes dissimilarity between expert and learner occupancy measures, and expert demonstration transformation and sharing reduce how often this happens based on absolute positions.

Transforming expert demonstration to generate informed and explainable multi-agent interactions improves learning representation and facilitates accurate behavior reconstruction. Furthermore, using $n$ individual discriminators while allowing them to share the transformed demonstrations reduces the complexity of each discriminator network and guarantees learners will be intuitively rewarded for all valid expert behavior. Nevertheless, these improvements introduce certain limitations. First, cohesion may not sufficiently describe multi-agent interactions in complex swarm scenarios involving multiple collective behaviors, and the search for a suitable interaction quantifier in such scenarios might not be trivial. A straightforward approach to address this is to transform expert demonstrations using several interaction metrics as in \cite{alharthi_automatic_2022}. However, this may introduce redundancy and increase the computation budget as the swarm size increases. Second, using $n$ discriminators and sharing expert demonstrations between them can introduce scalability issues as $n$ becomes very large. While we address this through distributed discriminator training in this work, several discriminators (e.g., $n > 100$) may be computationally expensive to train even in parallel. Finally, the challenge of collecting an adequate number of expert demonstrations remains, and we have not optimized our model to use minimal demonstrations. We intend to focus on these limitations in future works.

\section{CONCLUSION}
In this work, we reconstructed collective behavior in shared environments without having access to the swarm controller. We achieve this by transforming expert demonstrations into state features that sufficiently describe the multi-agent interactions between entities in the shared environments. We investigate three distinct classical swarm robotics. To improve learning representation, we separate the shared environments into motion and control layers and model the control layers as DEC-POMDPs grid world environments. Our results in the control layer demonstrate the capability of our MA-GAIL approach to accurately reconstruct observed collective behaviors in spatial organisation (i.e., aggregation, goal homing, and obstacle avoidance) compared to existing reconstruction algorithms. We show that transforming expert demonstrations into shared state features that sufficiently describe multi-agent interactions improves behavior reconstruction accuracy in distinct or unrelated swarm scenarios. In the future, we will investigate behavior reconstruction and recognition in complex practical scenarios involving multiple collective behaviors. As cohesion does not sufficiently describe all multi-robot interactions in more complex scenarios, we will investigate the automated discovery of explainable interaction quantifiers to achieve this. Future work will also consider the real-world constraints in experiments with physical multi-robot systems.

\bibliographystyle{IEEEtran}
\bibliography{IEEEabrv,iros24}

\end{document}